\documentclass[10pt,twocolumn,letterpaper]{article}
\usepackage[pagenumbers]{cvpr} % To force page numbers, e.g. for an arXiv version

\usepackage{graphicx}
\usepackage{amsmath}
\usepackage{amssymb}
\usepackage{booktabs}
\usepackage{lipsum}
\usepackage[toc,page]{appendix}
\usepackage{listings}
\usepackage{multirow}

\usepackage{fancyvrb}
\usepackage{listings}
\usepackage{color}
\usepackage{newtxtext, newtxmath}

\usepackage{algorithm2e}
\DontPrintSemicolon
\RestyleAlgo{ruled}
\SetKwComment{Comment}{// }{}

\definecolor{dkgreen}{rgb}{0,0.6,0}
\definecolor{gray}{rgb}{0.5,0.5,0.5}
\definecolor{mauve}{rgb}{0.58,0,0.82}

\usepackage[pagebackref,breaklinks,colorlinks]{hyperref}

\usepackage[capitalize]{cleveref}
\crefname{section}{Sec.}{Secs.}
\Crefname{section}{Section}{Sections}
\Crefname{table}{Table}{Tables}
\crefname{table}{Tab.}{Tabs.}

\def\cvprPaperID{5430} % *** Enter the CVPR Paper ID here
\def\confName{CVPR}
\def\confYear{2023}

\begin{document}

\title{3D Shape Augmentation with Content-Aware Shape Resizing}

\author{
Mingxiang Chen, Jian Zhang, Boli Zhou, Yang Song \\
Ant Group \\
% No. 7 Xinxi Street, Haidian District \\
Xihu District, Hangzhou, China \\
{\tt\small \{mingxiang.cmx, zj134362, zhouboli.zbl, zhaoshan.sy\}@antgroup.com} \\
% \and
}

\twocolumn[{%
\renewcommand\twocolumn[1][]{#1}%
\maketitle

\maketitle

\begin{center}
    \centering
    \captionsetup{type=figure}
    \includegraphics[width=.999\textwidth, height=.56\textwidth]{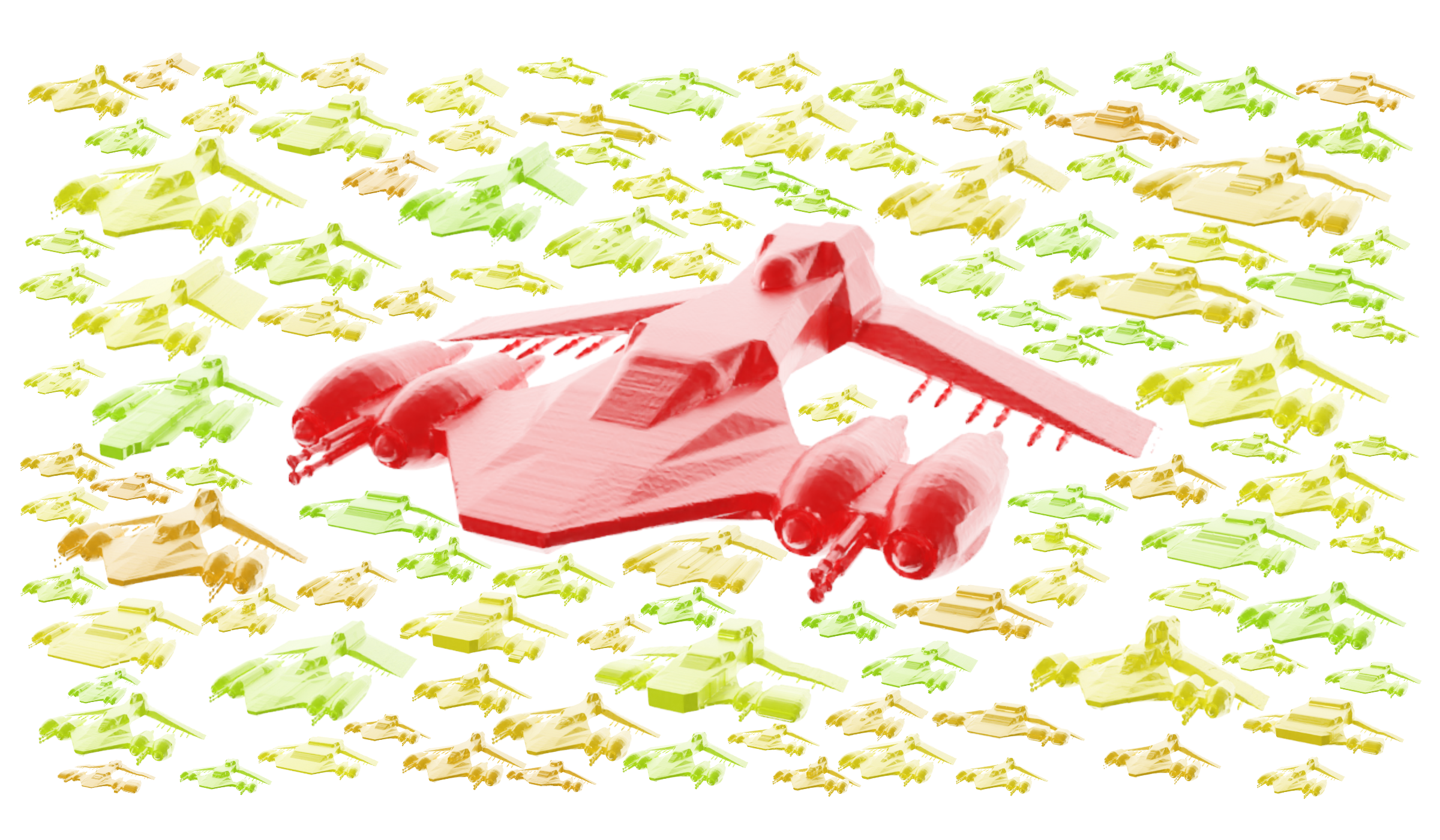}
    \captionof{figure}{Given an input model (rendered in \textbf{red} in the \textbf{middle}), our approach has the capability to generate a variety of augmented 3D shapes characterized by intricate structures and precise details.}
\end{center}%
}]

\begin{abstract}

Recent advancements in deep learning for 3D models have propelled breakthroughs in generation, detection, and scene understanding. However, the effectiveness of these algorithms hinges on large training datasets. We address the challenge by introducing Efficient 3D Seam Carving (E3SC), a novel 3D model augmentation method based on seam carving, which progressively deforms only part of the input model while ensuring the overall semantics are unchanged. Experiments show that our approach is capable of producing diverse and high-quality augmented 3D shapes across various types and styles of input models, achieving considerable improvements over previous methods. Quantitative evaluations demonstrate that our method effectively enhances the novelty and quality of shapes generated by other subsequent 3D generation algorithms.

\end{abstract}

\section{Introduction}

In recent years, there has been a rapid advancement in deep learning technologies associated with three-dimensional models. The corresponding algorithms have achieved significant breakthroughs in various tasks, including the generation, recognition, detection, and understanding of three-dimensional models and scenes. Despite the notable superiority of deep learning algorithms over traditional machine learning methods, their effectiveness hinges on the availability of large datasets for training. To address this challenge, the adoption of data augmentation has become widespread, especially in domains where substantial datasets are not readily accessible— a circumstance frequently encountered in the context of many 3D tasks. The primary objective of data augmentation is to generate supplementary data for training, and its efficacy in enhancing performance has been empirically validated.

A traditional and straightforward way of 3D shape augmentation is axis scaling, yet it is not content-aware, and the scaled 3D shape may be distorted. Alternative methods have also indicated limited effectiveness in addressing the issue of non-logical deformation. Though widely used in 2D image generation, detection, recognition, and many other tasks, the augmentation of 3D models has been relatively underexplored. On the other hand, the significance of varied and high-quality 3D content is growing within multiple industries, encompassing gaming, robotics, architecture, and social platforms. Constricting training datasets to a narrow scope of 3D models poses limitations on the algorithm's potential in various respects. Consequently, the development of an augmentation technique capable of generating diverse and high-quality 3D models is imperative.

In this paper, we present a novel yet straightforward augmentation method that produces diverse variations when given a 3D model as input. Our approach leverages the content-aware 2D image resizing technique based on seam carving, ensuring precise 3D seam prediction and enhanced computational efficiency. Additionally, we mitigate the issue of diversity by introducing the 'anchor points' into our approach.

Overall, our contributions are summarized as follows:

\begin{itemize}

\item We propose Efficient 3D Seam Carving, a content-aware 3D model augmentation algorithm, that evaluates which parts of the model can be deformed along which specific directions, thereby appropriately implementing different deformation strategies for various regions of an input model.

\item We introduce beam search and anchor point selection techniques to ensure that our method computes efficiently and outputs various 3D shapes.

\item We show compelling results and compare our method with previous methods. The results show that our method produces high-quality and diverse augmented 3D shapes among varied types of input models.

\end{itemize}

The paper is organized as follows. First, we introduce related works in Section \ref{sec: related works}. The details of our method are explained in Section \ref{sec: method}. The settings of experiments and their results are discussed in Section \ref{sec: exp}. The conclusion is presented in Section \ref{sec: conclusion}.

\section{Related Works} \label{sec: related works}

\textbf{Content-aware 2D image retargeting.}
Content-Aware Image Retargeting (CAIR) techniques are important for displaying images or videos on various devices with diverse aspect ratios \cite{asheghi2022comprehensive}. Compared to naive resizing methods such as uniform scaling and fixed-window cropping, CAIR methods preserve crucial regions of the input image while minimizing the occurrence of artifacts and distortion. Generally speaking, the CAIR methods can be classified into 4 categories \cite{asheghi2022comprehensive}: 1) discrete methods \cite{avidan2023seam, rubinstein2008improved, achanta2009saliency, hashemzadeh2019content}, 2) continuous methods \cite{gal2006feature, liang2012patchwise, du2013stretchability}, 3) multi-operator methods \cite{rubinstein2009multi, dong2009optimized}, and 4) deep learning based methods \cite{song2018carvingnet, liu2018composing, lin2019deepir}. Numerous methodologies have been proposed within each category, each with its own set of advantages and limitations. Among them, the Seam Carving (SC) \cite{avidan2023seam} algorithm is a set of discrete methods upon which many CAIR techniques are based. The SC method first computes the energy of each pixel based on a pre-defined energy function, typically indicating the pixel's significance. Subsequently, the algorithm identifies a low-energy seam and, based on the target dimensions, determines whether the seam is to be removed or inserted.

\textbf{Image and 3D shape augmentation.}
Data augmentation is a technique commonly used in computer vision and machine learning to artificially increase the size of a training dataset and avoid overfitting by applying various transformations to existing data. Image augmentation proves to be necessary for numerous 2D image processing tasks \cite{xu2023comprehensive, shorten2019survey}, including but not limited to segmentation \cite{ji2019invariant, long2015fully}, detection \cite{chen2023parsing, redmon2016you, girshick2015fast}, unsupervised learning \cite{he2020momentum, chen2021augnet, haeusser2019associative, ji2019invariant}, and various other applications. The 2D image augmentation techniques can be grouped into three main categories \cite{xu2023comprehensive}: model-free approaches, model-based approaches, and optimizing policy-based. The model-free approach leverages traditional image processing techniques such as transformations like cropping, rotation, flipping, and scaling, whereas the model-based approach capitalizes on image generation models \cite{yi2019generative, frid2018gan} to produce synthetic images. On the other hand, the optimizing policy-based method seeks to balance and find the most advantageous combination of both \cite{fawzi2016adaptive}. Moreover, image augmentation techniques may vary depending on the target domain, for example, medical images \cite{litjens2017survey, yi2019generative, frid2018gan}, agricultural images \cite{xu2022style}, and satellite images \cite{mohanty2020deep}.

As for the 3D shapes, to the best of our knowledge, random scaling \cite{nash2020polygen, siddiqui2023meshgpt} is probably the most widely adopted augmentation approach. Other methods include piecewise warping \cite{nash2020polygen}, shape uniting \cite{zheng2023locally}, spectral augmentation \cite{richardson2023texture}, and planar decimation \cite{nash2020polygen, siddiqui2023meshgpt} (as documented in Blender's manual \footnote{\url{https://docs.blender.org/manual/en/latest/modeling/modifiers/generate/decimate.html}}). These methodologies lack content-awareness. Consequently, the execution of major augmentation, such as a large scaling factor, may yield results with noticeable artifacts and distortions (except for planar decimation). Conversely, when minor augmentation is applied, the augmented model may differ imperceptibly from the original model, thereby potentially being detrimental to the algorithm's generalizability. In the context of 2D images, cropping serves as a straightforward remedy for the aforementioned distortion issues. However, in the case of 3D models, integrity is important, and thus, cropping is generally not considered a viable augmentation technique.

\section{Method} \label{sec: method}

\subsection{Overview}

A 3D shape can be represented as a discrete occupancy function $o:z \in \mathcal{Z} \mapsto {0,1}$ which is defined on a 3D grid $\mathcal{Z}$. $o(z)$ records the value of occupancy that $o(z)=1$ if the center of the grid cell is inside of the 3D shape and $o(z)=0$ otherwise. Let $\mathbf{G_{o}}$ be an $N_i \times N_j \times N_k$ grid containing the values of occupancies given a water-tight 3D shape. A spatial seam is defined to be
\begin{equation}
\mathbf{s^{z}} = \{s^{z}_{i,j}\} = \{(i,j,z(i,j))\}_{i,j},
\end{equation}
where $i,j,k$ are coordinate values on the x, y, and z axis, respectively. $i,j,k \in \mathbb{Z}$, $1 \leq i \leq N_i$, $1 \leq j \leq N_j$, $1 \leq k \leq N_k$, and $\forall i,j$, $|z(i,j)-z(i-1,j)| \leq 1$, $|z(i,j)-z(i,j-1)| \leq 1$. Here, $z$ is a mapping $z:[1,...,N_i] \times [1,...,N_j] \mapsto [1,...,N_k]$ representing the coordinate values on the $z$ axis. That is, a plane splits the 3D shape to two, from top to bottom, from left to right, and containing exactly one voxel for every single coordinate $(i,j)$. In this example, the x-axis and y-axis are associated with independent variables, while the z-axis is associated with the dependent variables. Spatial seams that cut the 3D shape from different directions (\eg from top to bottom, from front to back) are defined in similar ways.

Similar to the occupancy function, a discrete signed distance function (SDF) $g:z \in \mathcal{Z} \mapsto \mathbb{R}$ is defined on a 3D grid $\mathcal{Z}$, where the value records the signed distance from the center of the grid cell to the surface of the 3D shape. Specifically, we can restrict the values of the signed distance within a defined range, referred to as the Truncated Signed Distance Function (TSDF). The grid containing the SDF and TSDF values are denoted as $\mathbf{G_s}$ and $\mathbf{G_t}$, respectively. The definition of the spatial seam in the signed distance grid is analogous to the definition in the occupancy grid.

Given an energy function $e$, the cost of a spatial seam is
\begin{equation}
    E(s) = \sum_{i,j} e(\mathbf{G}(s^{z}_{i,j})),
\end{equation}
where we define the energy function as:
\begin{equation}
    e_z(\mathbf{G}) = |\frac{\partial}{\partial z} \mathbf{G}|,
    \label{eqn:e_z}
\end{equation}
and the variable $z$ in Eq. \eqref{eqn:e_z} changes following the direction of the spatial seam. We name it as the \textit{cutting axis}. Different from the energy function designed for 2D images \cite{rubinstein2008improved, avidan2023seam}, where the energy function is commonly associated with the first-order derivatives of the image along the horizontal and vertical axes, typically represented by the sum of their absolute values, we mainly focus on capturing the gradient information of the model exclusively along a specific axis. The rationale behind this observation stems from the fact that when the gradient along a specific axis of a 3D shape reaches zero, it commonly signifies the potential for scale variation along that particular axis. To illustrate, take a handleless cup where the gradients along the y-axis (representing the height direction) are zero in the central segment. If we elongate the central segment along the y-axis, we would achieve a slightly taller or shorter cup while preserving a common shape. In contrast, stretching the cup along the x-axis or z-axis would result in an elliptical cup, a less frequent occurrence. 

\begin{equation}
    e_3(\mathbf{G}) = |\frac{\partial}{\partial x} \mathbf{G}| + |\frac{\partial}{\partial y} \mathbf{G}| + |\frac{\partial}{\partial z} \mathbf{G}|.
    \label{eqn:e_full}
\end{equation}

\begin{figure}[!b]
    \centering
    \includegraphics[width=.99\linewidth]{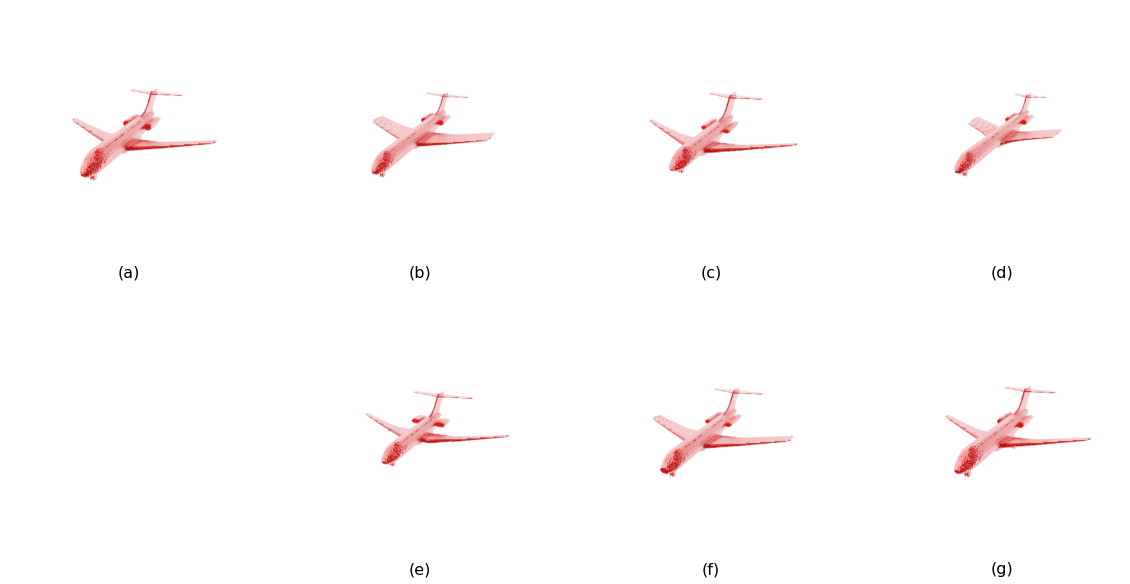}
    \caption{Comparison between different energy functions. Given the same input model (a), we show three augmented results (b-d) using Eq. \ref{eqn:e_z}, and another three results (e-g) using Eq. \ref{eqn:e_full}. The anchors are not limited to occupied cells in both cases.}\label{Fig:full_grad}
\end{figure}

In Section \ref{sec: exp}, a comparative analysis will be conducted between the aforementioned energy function and an alternative energy function considering the first-order derivatives along all three axes. In short, aside from the higher computational cost of using \ref{eqn:e_full}, the results of the two are similar. Our augmentation algorithm follows the structure outlined in Algorithm \ref{alg:overview}. Notably, the augmentation of a 3D shape involves adjustments along its three axes, where the model is rotated to designate its x-axis, y-axis, and z-axis successively as the \textit{cutting axis}. Furthermore, a maximum scaling factor denoted as $S_{max}$ ensures that the dimension $N$ of the model along a specific axis remains within the range $[N \times (1-S_{max}), N \times (1+S_{max})]$ after the augmentation.

\begin{algorithm}
% \linespread{1.15}\selectfont
\caption{Algorithm Overview}\label{alg:overview}
\textbf{Initialization:} $\mathbf{G}$, $n$ \Comment*[l]{beam size}
\ForEach{$axis \in \{\textbf{x}, \textbf{y}, \textbf{z}\}$}{
    $\mathbf{G} \gets$ Rotate($\mathbf{G}, axis$) \\
    $N_i, N_j, N_k \gets$ Shape($\mathbf{G}$) \\
    $delta \gets$ RandInt(-$N_k * S_{max}$, $N_k * S_{max}$+1) \\
    \For{$i = 0;\ i < delta;\ i = i + 1$}{
        $e \gets $ Eq. \eqref{eqn:e_z} \\
        $e_x \gets $ Eq. \eqref{eqn:e_2dx} \\
        $e_y \gets $ Eq. \eqref{eqn:e_2dy} \\
        $a \gets$ SetAnchor($e$) \\
        $p^x \gets$ BeamSearch2D($n, e_x, a$) \\
        $p^y \gets$ BeamSearch2D($n, e_y, a$) \\
        $seam^x \gets$ BeamSearch3D($n, e, p^x$) \\
        $seam^y \gets$ BeamSearch3D($n, e, p^y$) \\
        $seam \gets$ MinCostSeam($seam^x, seam^y$) \\
        \eIf{$delta > 0$}{
            $\mathbf{G} \gets$ SeamInsertion($\mathbf{G}, seam$)
        }{
            $\mathbf{G} \gets$ SeamRemoval($\mathbf{G}, seam$)
        }
    }
    $\mathbf{G} \gets$ RotateInverse($\mathbf{G}, axis$) \\
}
\end{algorithm}

% Nevertheless, removing the seam with the lowest energy fails to acknowledge the energy introduced by the emergence of new edges formed when previously disconnected voxels are brought into proximity following the removal of the seam. Hence, the forward energy \cite{rubinstein2008improved} is used that the seam removal inserts minimal amount of $\hat{e}_z(\mathbf{G})$ which is as follows

% \begin{equation}
%     e_z(i,j,k) = \hat{e}_z
%     \label{eqn:e_z}
% \end{equation}

\subsection{Beam Search}

Solving the global optimal energy or minimal cut on a graph based on a 2D image or 3D video volume is simply not feasible \cite{rubinstein2008improved}. Hence, the beam search algorithm is used to improve efficiency. This effective heuristic search algorithm enables the generation of coherent sequences by exploring multiple candidate hypotheses simultaneously. Firstly, an energy map is initialized using Eq. \eqref{eqn:e_z}. Subsequently, a 2D energy map is obtained by reducing the original map on the x or y axis, which is named as the \textit{reducing axis}:

\begin{equation}
    e_{x, 2D}(j,k) = \sum_i e_z(i,j,k),
    \label{eqn:e_2dx}
\end{equation}

\begin{equation}
    e_{y, 2D}(i,k) = \sum_j e_z(i,j,k).
    \label{eqn:e_2dy}
\end{equation}
The remaining axis, which is neither the \textit{cutting axis} nor the \textit{reducing axis}, is hereby referred to as the \textit{main axis}. Given an initial anchor point on the 2D energy map, three possible neighbors on top or bottom (along the \textit{main axis}) of it are extended. These extensions are then scored based on the sum of their energy. Nevertheless, the cumulative inclusion of the three neighbors above or below the current cell into the existing path will result in an exponential increase in the overall number of plausible paths. Hence, pruning is then applied to retain only the top-n candidate extensions, effectively reducing the search space. The algorithm iteratively repeats the expansion, scoring, and pruning steps until the candidate paths connect the pixels from the top to the bottom of the 2D energy map. Finally, the lowest-cost path is selected as the anchor path. Then the aforementioned procedure is repeated on the 3D energy map, with the only distinction being that the initial anchor, which was a single point, is now replaced by the anchor path obtained from the 2D energy map.

\begin{algorithm}
\caption{Beam Search on the 2D Energy Map}\label{alg:beam2d}
\textbf{Initialization:} $n$, $e_{2D}$, $a$\\
$(x_a,y_a) \gets a$ \Comment*[l]{anchor point}
$h, w \gets$ Shape($e_{2D}$) \\
$p \gets$ list($h$) \Comment*[l]{$p$ is a list of length $h$.}
$p[x_a] \gets y_a$\;
$beam \gets [p]$\;
\eIf{$x_a = 1$}{
    $x \gets x_a + 1$\;
}{
    $x \gets x_a - 1$\;
}
\While{$x < h$}{
    $b \gets []$\;
    \ForEach{$p \in beam$}{
        % \eIf{$x < x_a$}{
        %     $x_{prev} \gets x + 1$ \\
        % }{
        %     $x_{prev} \gets x - 1$ \\
        % }
        % $y_{prev} \gets p[x_{prev}]$ \\
        % $ys \gets [y_{prev} - 1, y_{prev}, y_{prev} + 1]$ \\
        \eIf{$x < x_a$}{
            $ys \gets [p[x + 1] - 1, p[x + 1], p[x + 1] + 1]$ \\
        }{
            
            $ys \gets [p[x - 1] - 1, p[x - 1], p[x - 1] + 1]$ \\
        }
        \ForEach{$y \in ys$}{
            $p_{new} \gets$ copy($p$) \\
            $p_{new}[x] \gets y$ \\
            $b$.append($p_{new}$) \\
        }
        $beam \gets$ top$_n(b, n, e_{2D})$ \\
    }
    
    \uIf{$x = 1$}{
        $x \gets x_a + 1$\;
    }
    \uElseIf{$x < x_a$}{
        $x \gets x - 1$\;
    }
    \Else{
        $x \gets x + 1$\;
    }
}
\Return beam[0]
\end{algorithm}

% \begin{algorithm}
% \caption{Beam Search on the 3D energy grid}\label{alg:beam3d}
% \KwData{$n \geq 0$}
% \KwResult{$y = x^n$}
% $y \gets 1$\;
% $X \gets x$\;
% $N \gets n$\;

% \While{$N \neq 0$}{
%     \eIf{$N$ is even}{
%         $X \gets X \times X$\;
%         $N \gets \frac{N}{2}$ \Comment*[r]{This is a comment}
%     }{
%         \If{$N$ is odd}{
%             $y \gets y \times X$\;
%             $N \gets N - 1$\;
%         }
%     }
% }
% \end{algorithm}

\subsection{Diversity}

\begin{figure}[!tb]
    \centering
    \includegraphics[width=.99\linewidth]{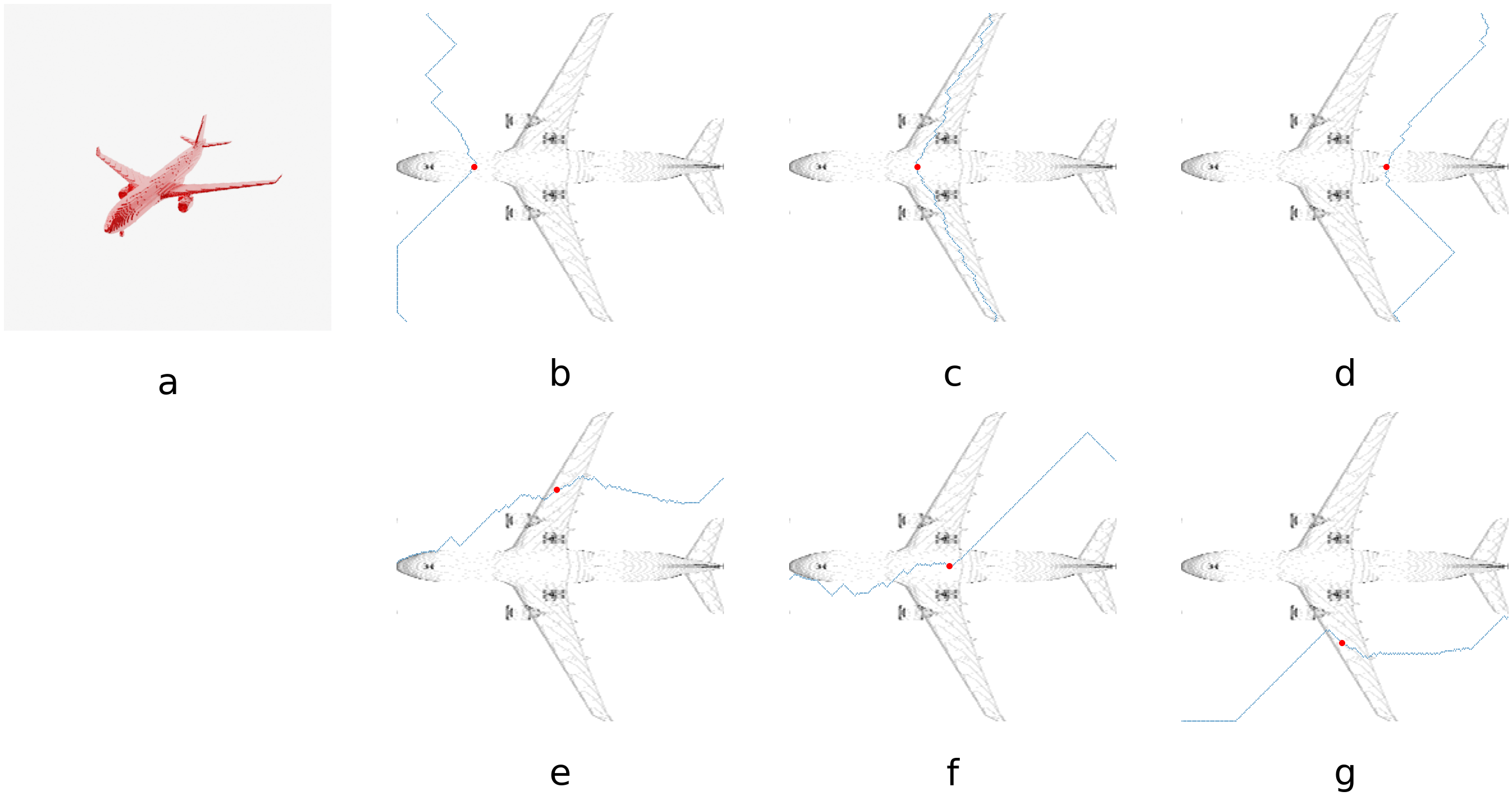}
    \caption{The selection of the anchor point can notably impact the trajectory of the seam. Taking an occupancy grid of a plane as an example (a), the energy maps are reduced along the y-axis (b-g). The anchors (illustrated as red dots) vary along both the z-axis (b-d) and x-axis (e-g), and the \textit{cutting axis} also transitions from the z-axis (b-d) to the x-axis (e-g). Both modifications contribute to a substantial divergence in the seams' paths.}\label{Fig:choice_of_anchor}
\end{figure}

\begin{figure}[!b]
    \centering
    \begin{subfigure}[t]{0.24\linewidth}
        \includegraphics[width=\linewidth]{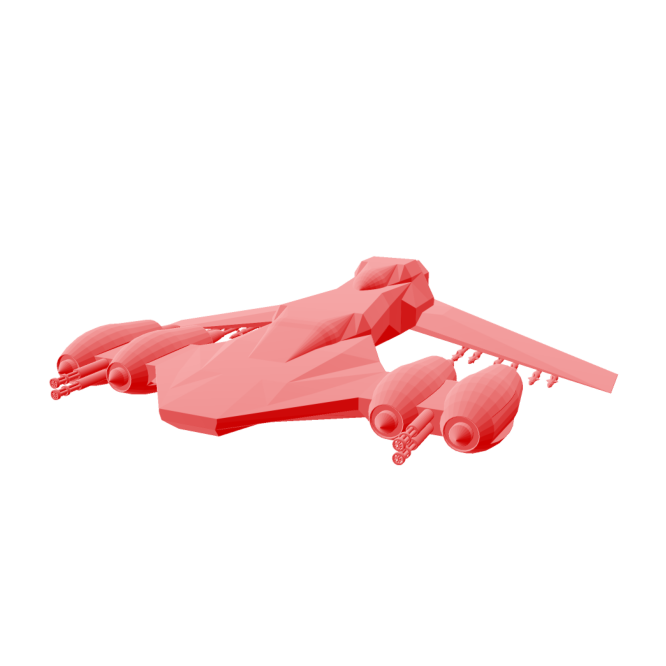}\label{fig:result_on_axis_0}
        % \caption{Pok\'emon}
    \end{subfigure}
    \begin{subfigure}[t]{0.24\linewidth}
        \includegraphics[width=\linewidth]{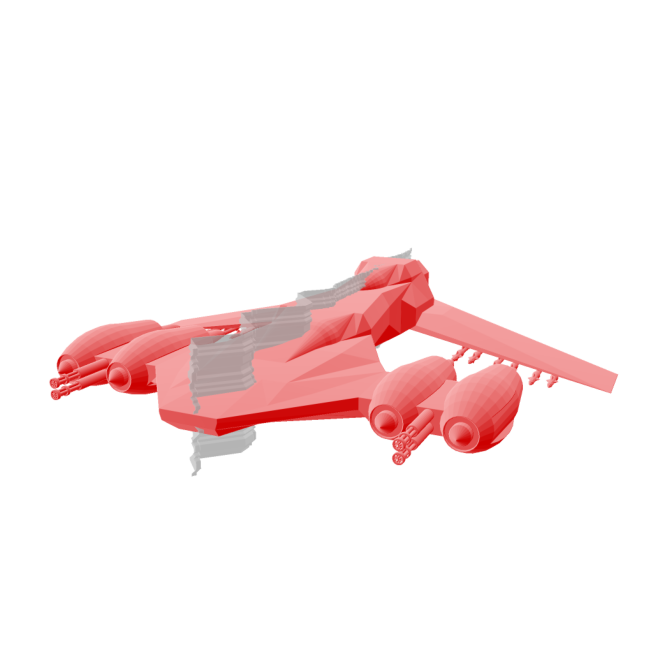}\label{fig:result_on_axis_1}
        % \caption{Pok\'emon}
    \end{subfigure}
    \begin{subfigure}[t]{0.24\linewidth}
        \includegraphics[width=\linewidth]{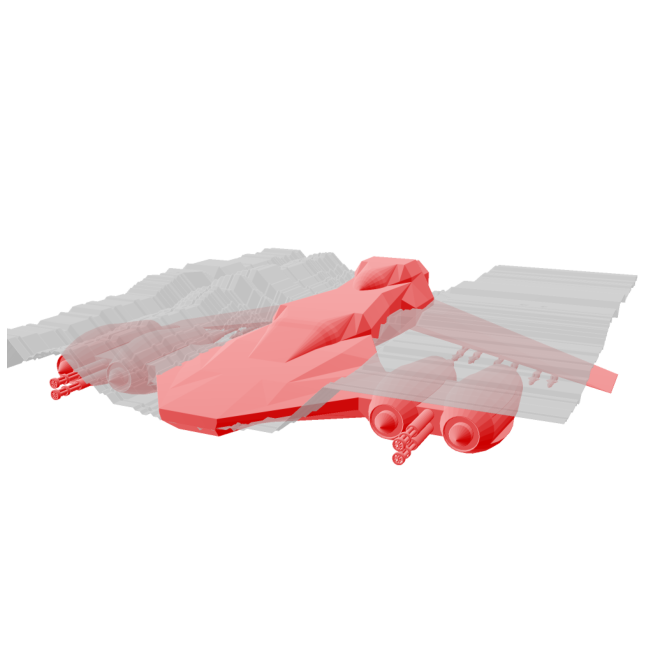}\label{fig:result_on_axis_2}
        % \caption{Pok\'emon}
    \end{subfigure}
    \begin{subfigure}[t]{0.24\linewidth}
        \includegraphics[width=\linewidth]{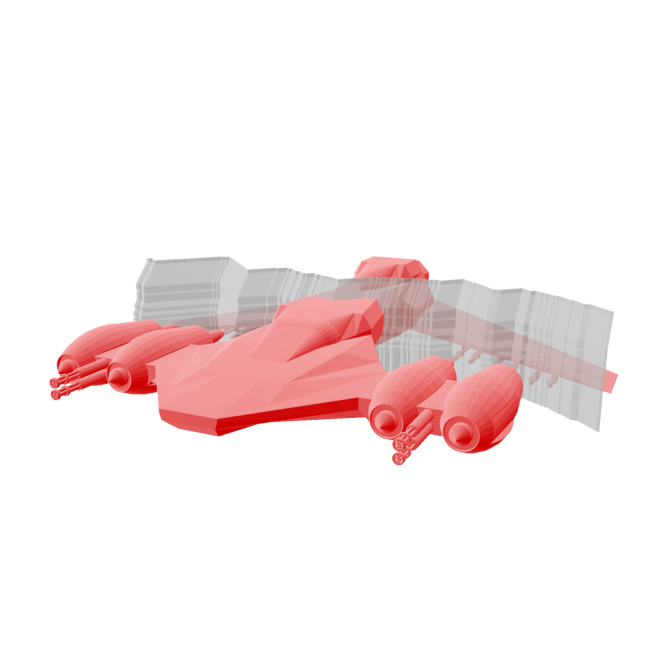}\label{fig:result_on_axis_3}
        % \caption{Pok\'emon}
    \end{subfigure}
    
    \caption{An illustration of seams computed using our algorithm, with different \textit{cutting axis} employed as input.}\label{Fig:result_on_axis}
\end{figure}

Beyond considerations of computational complexity, an additional rationale for not using global optimal energy lies in our expectation for each augmentation to yield distinct 3D shapes. As depicted in Fig. \ref{Fig:choice_of_anchor}, the selection of anchor points and the orientation of spatial seams exert a substantial influence on the trajectory of the seam, thereby ensuring diversity in the resulting models after each augmentation.

Note that for a relatively regular 3D shape, the partial derivatives of its voxels along a particular axis are mostly zero or small values. Therefore, when selecting an anchor point from a 2D energy map, a naive approach is to randomly choose a pixel as the anchor. However, as the number of iterative steps increases, the outcomes of augmentations may manifest only marginal variations. To address this, the possible anchor locations are firstly partitioned into several clusters using mini-batch $k$-means \cite{sculley2010web}. Subsequently, top $k$/3 plausible clusters are selected, ranked by their average cost across $s$ simulations on the 2D energy map, with random points within these clusters designated as anchors. At the beginning of each augmentation, $m$ clusters are selected from the top $k$/3 clusters. For each subsequent step within the augmentation, a cluster is chosen from the selected subset, and a random point within the chosen cluster is assigned as the anchor. In our method, the possible anchors are defined as the cells that are both occupied (or in other words, inside the 3D shape) and the corresponding energy on $e_{z}$ is less than a threshold $\epsilon$, which equals $1 \times 10 ^ {-3}$ in the experiments below.

In contrast to seam carving in images, for an occupancy grid or TSDF grid, the beam search process yields numerous candidate paths with equivalent cumulative gradients due to non-zero gradients occurring only on or near the model's surface during the search. To ensure that beam search can maximally expand the search space, thereby finding solutions closer to the optimal, the beam search is designed to maintain solutions with larger mutual distances when multiple paths have equal costs during the seam search. More specifically, consider the following scenario: suppose we have 12 paths and the beam size equals to 4. 1) If exactly 4 paths have a cumulative gradient less than a specific threshold T, then retain those 4 paths; 2) If there exists a set A containing three paths, where the cumulative gradient of each path is less than the threshold T, and another set B containing three paths, where the cumulative gradient of each equals T, and the cumulative gradients of the remaining six paths are greater than T, then select a path from B with the smallest total distance to all paths in B and place it into set A, and retain the paths in set A; 3) If, in the second case, the difference between the number of paths in set A and the beam size is greater than 1, then first select a path from B with the largest total distance to all paths in B and place it into a candidate set C, and continuously select the path from B that has not been placed into C and also has the largest minimum distance to all paths in C until the sum of the sizes of A and C equals the beam size. Finally, retain the paths in set A and set C. The distance between two paths $p_1$ and $p_2$ is defined as:

\begin{equation}
    D(p_1, p_2) = \sum_x |p_1[x] - p_2[x]|.
    \label{eqn:path_dist}
\end{equation}

% \begin{algorithm}
% \caption{Set anchors with Kmeans}\label{alg:speccluster}
% \KwData{$n \geq 0$}
% \KwResult{$y = x^n$}
% $y \gets 1$\;
% $X \gets x$\;
% $N \gets n$\;

% \While{$N \neq 0$}{
%     \eIf{$N$ is even}{
%         $X \gets X \times X$\;
%         $N \gets \frac{N}{2}$ \Comment*[r]{This is a comment}
%     }{
%         \If{$N$ is odd}{
%             $y \gets y \times X$\;
%             $N \gets N - 1$\;
%         }
%     }
% }
% \end{algorithm}

\subsection{Symmetry Check}

Numerous methodologies are available for evaluating the symmetry of an input 3D shape. One approach involves examining the chamfer distance between sampled surface points from each half and comparing it with a predetermined threshold. In our proposed method, we quantify symmetry using the elementwise mismatch rate between two halves. Taking inspiration from the $F_1$ score, which appropriately balances precision and recall, we employ the harmonic mean of mismatch rates for both occupied and unoccupied cells:

\begin{equation}
m_{x}(i,j,k) = \mathbf{G_o}(i,j,k) \oplus \mathbf{G_o}(N_i-i,j,k)
\label{eqn:mx}
\end{equation}

% \begin{equation}
% m_{y} = \mathbf{G_o}(i,j,k) \oplus \mathbf{G_o}(i,N_j-j,k)
% \label{eqn:my}
% \end{equation}

% \begin{equation}
% m_{z} = \mathbf{G_o}(i,j,k) \oplus \mathbf{G_o}(i,j,N_k-k)
% \label{eqn:mz}
% \end{equation}

\begin{equation}
    rate_{o,x} = \frac{\sum_{i,j,k} m_{x}(i,j,k) \times \mathbf{G_o}(i,j,k)}{\sum_{i,j,k} \mathbf{G_o}(i,j,k)}
    \label{eqn:rox}
\end{equation}

\begin{equation}
    rate_{u,x} = \frac{\sum_{i,j,k} m_{x}(i,j,k) \times (1 - \mathbf{G_o}(i,j,k))}{\sum_{i,j,k} 1 - \mathbf{G_o}(i,j,k)}
    \label{eqn:rux}
\end{equation}

\begin{equation}
    rate_{x} = 2 * \frac{rate_{o,x} * rate_{u,x}}{rate_{o,x} + rate_{u,x}}
    \label{eqn:rx}
\end{equation}
where $\oplus$ stands for the XOR operation, and $rate_x$ represents the mismatch rate of the input 3D shape along the x-axis. Our definition of mismatch rate along the y-axis and z-axis is analogous to that of $rate_x$, a redundant elaboration is omitted at this juncture.

A 3D shape is considered symmetrical along a specific axis when the mismatch rate is below a predefined threshold, denoted as $T_s$. If the condition is satisfied, given that it would produce two results through mirror transformations (as long as the path is not symmetrical along that axis), the output is determined by selecting the seam with the lower total energy. If the input model is symmetrical along multiple axes, an exhaustive exploration of all feasible combinations of mirror transformations is conducted, and the optimal seam is selected as the final output.

\subsection{Performance Trade-offs}

Solving the near-optimal seam is hard, particularly when dealing with high-resolution grids. In our experiments, augmenting a 3D shape with a resolution under 128 may require a few seconds, whereas resolutions exceeding 512 can extend the process to several minutes. One plausible optimization involves solely executing a 2D beam search, and stacking the identified path along the \textit{reducing axis} to establish the resulting seam. Despite receiving significant performance improvement, this approach is accompanied by a substantial decline in seam quality. Hence, we introduce an additional filtering mechanism that if the average energy of the seam surpasses a threshold $T_c$, the seam is discarded. In our conducted experiments, the algorithm gives satisfactory results when:

\begin{equation}
e_{avg} = \frac{\sum_{i,j,k} e(i,j,k)}{N_i \times N_j \times N_k}
\label{eqn:grad}
\end{equation}

\begin{equation}
    T_c= 
\begin{cases}
    e_{avg} * 0.25, & \text{if } e_{avg} > 4 \times 10^{-4}\\
    1 \times 10^{-4}, & \text{otherwise}
\end{cases}
\label{eqn:tc}
\end{equation}

\section{Experiments} \label{sec: exp}

\textbf{Dataset.}
We employ ShapeNetV2 \cite{chang2015shapenet} as the evaluation set for our augmentation method. In our experiments, we use Open3D to calculate the TSDF grid and the occupancy grid, which integrates of multiple depth images to derive the TSDF representation. Unless specified, the models in the following experiments are represented by occupancy grids.

\textbf{Qualitative demonstration.}
The augmented outcomes by our method are compared with those obtained through axis scaling \cite{nash2020polygen}, piecewise linear warping \cite{nash2020polygen}, and spectral augmentation technique \cite{richardson2023texture}. In alignment with \cite{nash2020polygen}, we independently scale each axis by uniformly sampling scaling factors within the range $[0.75, 1.25]$. For the piecewise linear warping, we first normalize the vertices of the models to constrain the coordinate values to the interval $[-1, 1]$. Subsequently, we define a piecewise linear warping function by partitioning the interval [-1, 1] into 6 even sub-intervals. The warping factors are sampled from a log-normal distribution with a standard deviation of 0.25. Consistent with the previous research, the warping factors exhibit symmetry with respect to the x and z-axes. We adopt identical configurations as outlined in \cite{richardson2023texture} during the implementation of spectral augmentation.

As shown in Fig. \ref{Fig:full_grad}, we compare the augmented results using Eq. \ref{eqn:e_z} with an alternative function taking the first-order derivatives along all three axes into account using Eq. \ref{eqn:e_full}. In most cases, the results are similar. One difference is that if the anchor cell is not an occupied cell (which is not true in our method), the seam paths are sometimes likely to pass through the tips of the 3D model where the sum of the energy is low if Eq. \ref{eqn:e_full} is used. This can result in a 3D shape deviating from common expectations, for example, the flattened tips of the planes' noses.

\begin{figure}[!tb]
    \centering
    \includegraphics[width=.99\linewidth]{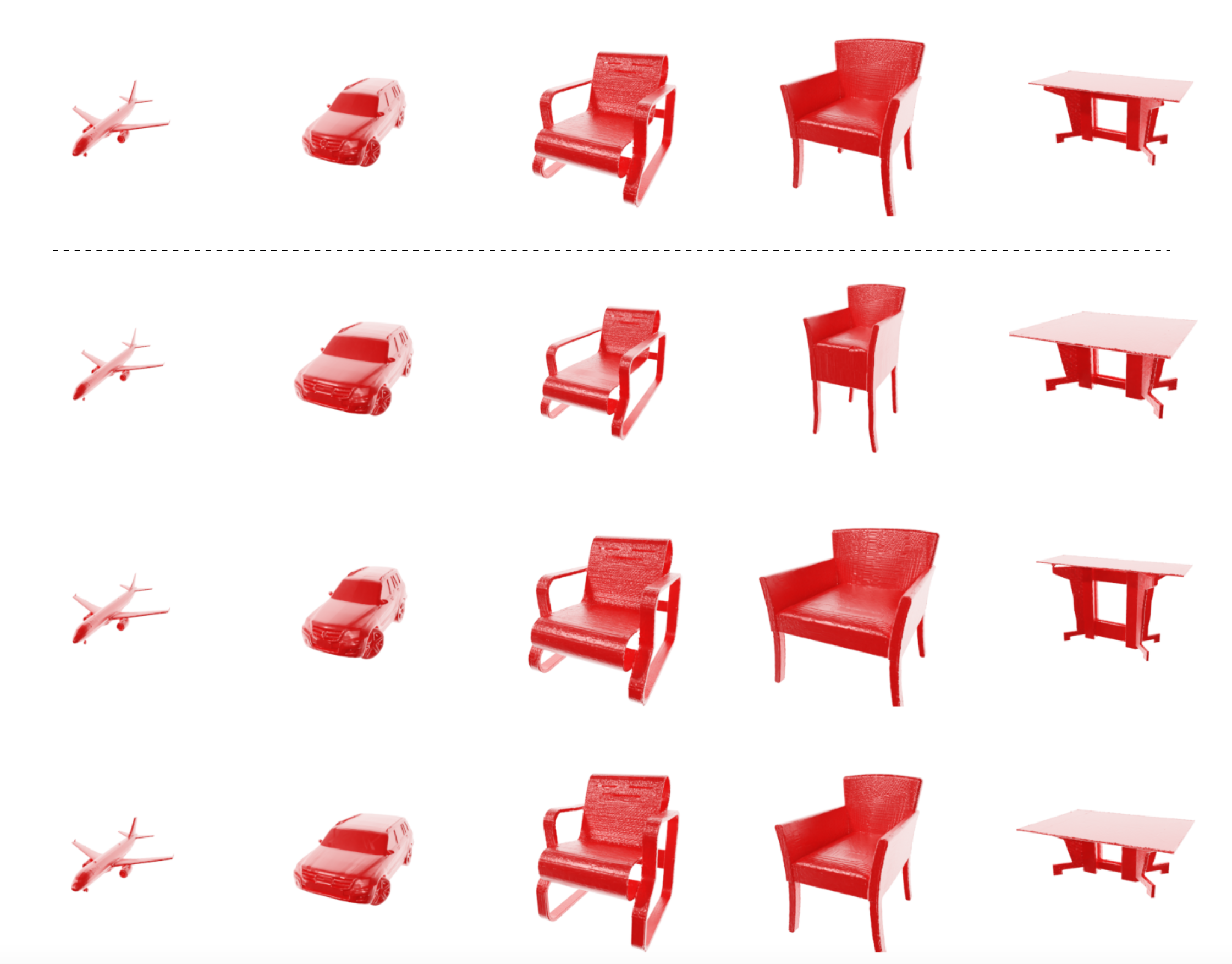}
    \caption{Illustration of augmented models based on input 3D shapes represented by TSDF grids. The first row of each column demonstrates the input models.}\label{Fig:sdf}
\end{figure}

\begin{figure*}[!tb]
    \centering
    \includegraphics[width=.999\linewidth, height=.5\linewidth]{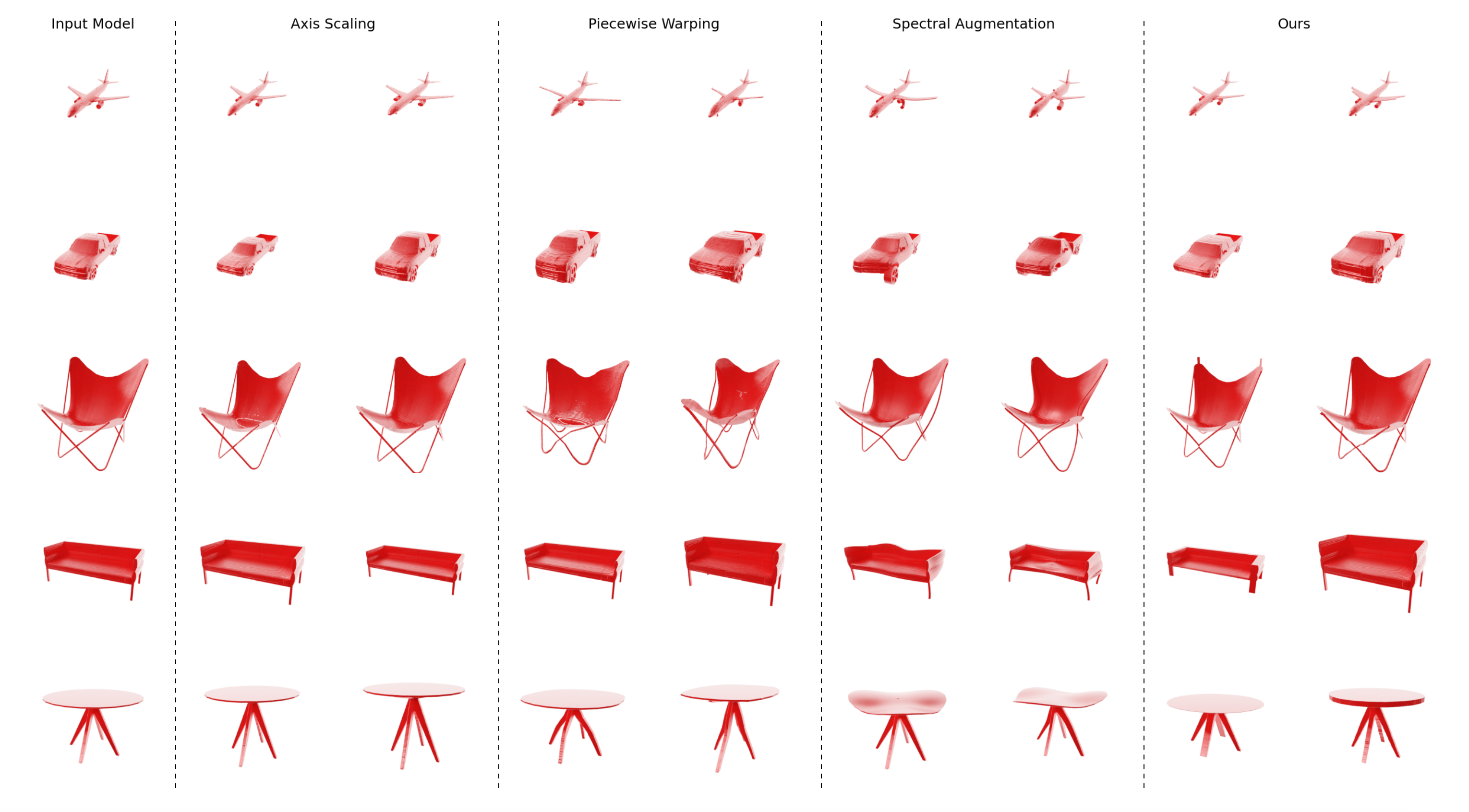}
    \caption{Qualitative comparison with axis scaling \cite{nash2020polygen}, piecewise warping \cite{nash2020polygen}, and spectral augmentation \cite{richardson2023texture}. We show the results obtained by augmenting each input model twice using the four methods, respectively.}\label{Fig:occ_illustration}
\end{figure*}

In Fig. \ref{Fig:occ_illustration} and Fig. \ref{Fig:sdf}, we present illustrative instances of augmented 3D shapes spanning various genres and styles. We utilize occupancy grids and TSDF grids to represent the models, respectively. The outcomes demonstrate the effectiveness of our approach in augmenting a variety of 3D shapes that align with common expectations.

\begin{table*}[!tb]
    \renewcommand{\arraystretch}{1.2}
    \centering
    \caption{Quantitative evaluation between the generated shapes produced by Neural Wavelet-domain Diffusion \cite{hui2022neural} with (\textbf{ours}) and without (\textbf{original}) our augmentation method. From the result, we can see that in most cases, our augmentation method enhances the quality of the outcome especially when the training data is insufficient.}
    \setlength{\tabcolsep}{1em}
    \resizebox{0.95\linewidth}{!}{
    % \scriptsize
    \begin{tabular}{ c | c c c c | c c c c | c c c c }
    \toprule
        & \multicolumn{4}{|c|}{MMD $\downarrow$} & \multicolumn{4}{|c}{COV $\uparrow$} & \multicolumn{4}{|c}{1-NNA $\sim0.5$} \\
        & \multicolumn{2}{|c}{CD} & \multicolumn{2}{c|}{EMD} & \multicolumn{2}{|c}{CD} & \multicolumn{2}{c}{EMD} & \multicolumn{2}{|c}{CD} & \multicolumn{2}{c}{EMD} \\
        & original & ours & original & ours & original & ours & original & ours & original & ours & original & ours \\
        \midrule
        Airplane (2000) & \textbf{0.0026} & 0.0029 & 0.0958 & \textbf{0.0954} & 0.4630 & \textbf{0.4635} & 0.3790 & \textbf{0.3840} & \textbf{0.7423} & 0.7605 & 0.7878 & \textbf{0.7680} \\
        \hline
        % Bus (929) & 0 & 0 & 0 & 0 & 0 & 0 & 0 & 0 & 0 & 0 & 0 & 0 \\
        % \hline
        Bookshelf (452) & \textbf{0.0087} & 0.0089 & 0.1445 & \textbf{0.1442} & 0.5243 & \textbf{0.5377} & \textbf{0.5288} & 0.4823 & 0.5409 & \textbf{0.4878} & 0.5420 & \textbf{0.5265} \\
        \hline
        % train (377) & 0 & 0 & 0 & 0 & 0 & 0 & 0 & 0 & 0 & 0 & 0 & 0 \\
        % \hline
        % piano (236) & 0 & 0 & 0 & 0 & 0 & 0 & 0 & 0 & 0 & 0 & 0 & 0 \\
        % \hline
        Bed (233) & 0.0296 & \textbf{0.0134} & 0.2581 & \textbf{0.1950} & 0.3090 & \textbf{0.5837} & 0.3605 & \textbf{0.5494} & 0.8541 & \textbf{0.4485} & 0.8476 & \textbf{0.5386} \\
        \hline
        Tower (132) & 0.0205 & \textbf{0.0098} & 0.1952 & \textbf{0.1613} & 0.3864 & \textbf{0.5227} & 0.3864 & \textbf{0.4697} & 0.7879 & \textbf{0.5568} & 0.7386 & \textbf{0.6326} \\
        \hline
        Camera (113) & 0.0428 & \textbf{0.0196} & 0.2743 & \textbf{0.2058} & 0.3540 & \textbf{0.6106} & 0.3186 & \textbf{0.6106} & 0.8186 & \textbf{0.3938} & 0.7655 & \textbf{0.4248} \\
    \toprule
    \end{tabular}
    }
    \label{tab:quant_eval}
\end{table*}

\begin{figure*}[!tb]
    \centering
    \includegraphics[width=.999\linewidth, height=.39\linewidth]{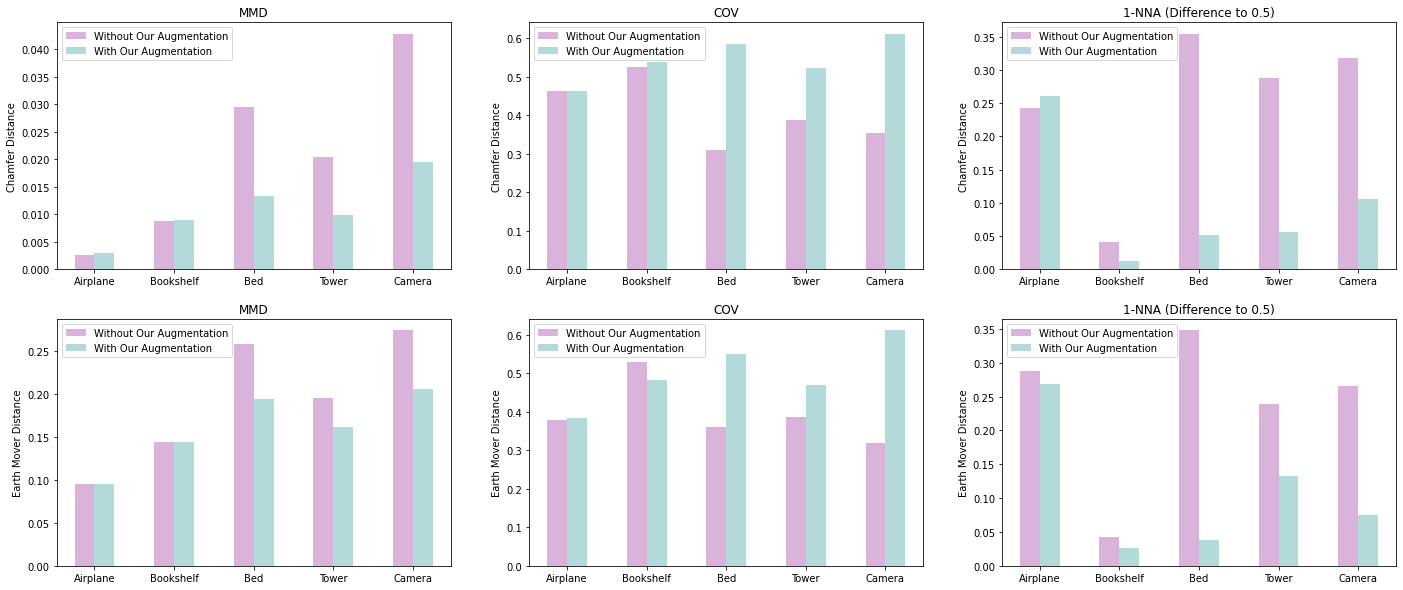}
    \caption{Qualitative assessments across various metrics reveal that our augmentation technique improves the outcomes' quality in most cases. To illustrate a more intuitive comparison of disparities, we computed the absolute difference between 1-NNA results and 0.5.}\label{Fig:metric_graph}
\end{figure*}

\textbf{Quantitative Evaluation.}
Intuitively, 3D model augmentation should enhance the quality and diversity of results generated by shape-generation algorithms. We employ Neural Wavelet-domain Diffusion \cite{hui2022neural} as the baseline model and compare the outcomes with and without our augmentation method during training. It is noteworthy that, apart from the airplane class models, which contain more than 4000 samples, the number of models in other categories is relatively limited. Specifically, the categories of bookshelf, bed, tower, and camera contain 452, 233, 132, and 113 models, respectively. Prior to training, we augment each model 8 times.

In line with the previous research \cite{hui2022neural, yang2019pointflow}, we uniformly sample 2,048 points on each generated shape and evaluate the shapes using three evaluation metrics: 1) minimum matching distance (MMD), which assesses the fidelity of the generated shapes; 2) coverage (COV), indicating the extent to which the generated shapes encompass the given 3D repository; and 3) 1-NN classifier accuracy (1-NNA), measuring the effectiveness of a classifier in distinguishing the generated shapes from those in the repository. Generally, a low MMD, a high COV, and a 1-NNA close to 50\% signify satisfactory generation quality. For the airplane models, we generated 2,000 models for evaluation purposes. For other categories, we generated an equal number of models as the training set for evaluation.

From Table \ref{tab:quant_eval}, it is evident that our augmentation method significantly enhances the performance of shape generation algorithms based on occupancy grid, SDF grid, or TSDF grid, particularly when the training data is insufficient.

\textbf{Human preference study.}
Given that one of the primary goals of our approach is to enlarge the 3D shape dataset, the augmented data holds potential for various downstream applications, such as 3D shape generation, classification, detection, and so on. Consequently, it is imperative for the augmented models to exhibit both logical coherence and diversity. Otherwise, the training set for downstream tasks may be contaminated, thereby yielding unsatisfying outcomes. Evaluating the quality of 3D shapes is a complex task. We focus on quantitative metrics derived from human preference evaluations, considering two key dimensions: diversity and visual quality.

Using models sourced from ShapeNetV2 \cite{chang2015shapenet} as our initial seed, we apply various methods to generate eight augmented 3D shapes for each model. A questionnaire is then presented, displaying eight 3D shapes augmented by our method compared to eight by a baseline. Both sets share identical rendering configurations and originate from the same source model. Subsequently, we pose two binary choice questions to assess preferences: 'Among the two groups of 3D models, which of them do you think makes more sense to you? In other words, which one do you think exhibits a higher quality?' and 'Which group do you think is more diverse? (Please exclude models that you consider to be of low quality when evaluating.)'. Our study involves 150 participants, each compares 10 pairs of results (2 models from each category).

As shown in Table \ref{tab:human_eval}, our proposed method shows competitive performance against axis scaling in terms of model quality and is more preferred compared to other baselines on the rest dimensions. It utilizes gradients to evaluate for which parts and in which directions the model can be deformed. This capability is likely the key factor contributing to its potential for achieving higher human preference.

\begin{table}
    \centering
    \caption{Quantitative evaluation on human preference of our proposed method over three baselines on diversity and quality.}
    \setlength{\tabcolsep}{1em}
    \resizebox{0.99\linewidth}{!}{
    \begin{tabular}{l c c }
    \toprule
        & Quality & Diversity \\
        \midrule
        Ours over axis scaling & 47.2\% & \textbf{79.4\%} \\
        Ours over axis piecewise warping & \textbf{78.6\%} & \textbf{63.6\%} \\
        Ours over axis spectral augmentation & \textbf{91.4\%} & \textbf{85.0\%} \\
    \toprule
    \end{tabular}
    }
    \label{tab:human_eval}
\end{table}

\begin{figure}
    \centering
    \includegraphics[width=.99\linewidth]{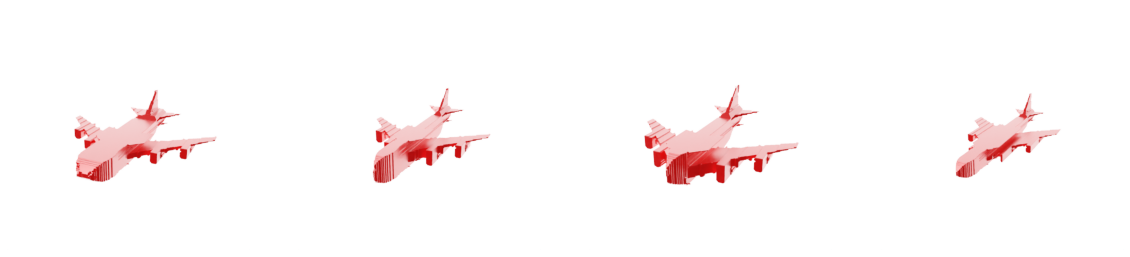}
    \caption{An illustration of our method fails to generate high-quality augmentations for certain 3D models.}\label{Fig:badcase}
\end{figure}

\textbf{Limitations.} Although the approach yields augmented 3D shapes of high quality, we observed that the generated shapes may exhibit artifacts when applied to pixel-style 3D shapes. Illustrated in Fig. \ref{Fig:badcase}, a genuine jet plane does not feature square engines or a square body, introducing irregular gradients. Our method struggles to accurately estimate the appropriate scaling direction for these components. In some other instances, our approach generates augmented desks and chairs with uncommon shapes. For instance, we observe tables with unusually chunky tabletops and chairs with very thick backrests.

\section{Conclusion} \label{sec: conclusion}

In this paper, we introduce a novel 3D model augmentation method based on 3D seam carving. Our approach produces diverse and high-quality results across various types of models in the form of both occupancy grids and SDF grids. The proposed method may be useful for many tasks related to 3D models such as model generation, detection, scene segmentation, 3D model recognition, and so on. We expect that this method could offer more pronounced assistance for training deep learning models with smaller datasets.

%%%%%%%%% REFERENCES

{\small
\bibliographystyle{ieee_fullname}
\bibliography{main}
}

% \newpage~

% \input{article/appendix}

\end{document}